\documentclass{article}

\usepackage[sort&compress, numbers]{natbib}
\usepackage[preprint]{neurips_2021}

\usepackage[utf8]{inputenc} % allow utf-8 input
\usepackage[T1]{fontenc}    % use 8-bit T1 fonts
\usepackage{hyperref}       % hyperlinks
\usepackage{url}            % simple URL typesetting
\usepackage{booktabs}       % professional-quality tables
\usepackage{amsfonts}       % blackboard math symbols
\usepackage{nicefrac}       % compact symbols for 1/2, etc.
\usepackage{microtype}      % microtypography
\usepackage{xcolor}         % colors
\usepackage{graphicx}
\usepackage{amsmath}
\usepackage{floatrow}

\title{ROPUST: Improving Robustness through Fine-tuning with Photonic Processors and Synthetic Gradients}

\author{%
  \textbf{Alessandro Cappelli}\thanks{Corresponding author: \texttt{alessandro@lighton.ai}} $^{\,1}$,
  \textbf{Julien Launay}$^{1,2}$,
  \textbf{Laurent Meunier}$^{3,4}$,
  \textbf{Ruben Ohana}$^{1,2}$,
  \textbf{Iacopo Poli}$^{1}$ \\
  $^{1}$LightOn, France \quad $^{2}$LPENS, École Normale Supérieure, France \\
  $^{3}$ Facebook AI Research, France \quad $^{4}$ Université Paris-Dauphine, France\\
}

\begin{document}

\maketitle

%submitted abstract: Adversarially robust deep learning models are typically learned through expensive adversarial training with Projected Gradient Descent. We introduce ROPUST, a remarkably simple and efficient method to leverage robust pre-trained models and further increase their robustness, at no cost in natural accuracy. Our technique relies on the use of an Optical Processing Unit (OPU), a photonic co-processor, and a fine-tuning step performed with direct feedback alignment, a synthetic gradient training scheme. We empirically test our method on six different models and three attacks in RobustBench, consistently improving over state-of-the-art performance. We also perform a phase retrieval attack, specifically designed to increase the threat level of attackers against our own defense, and show that its effect can be mitigated by taking simple precautions, even in an ideal setting without measurement noise and full access and knowledge of the system.

\begin{abstract}
Robustness to adversarial attacks is typically obtained through expensive adversarial training with Projected Gradient Descent. Here we introduce ROPUST, a remarkably simple and efficient method to leverage robust pre-trained models and further increase their robustness, at no cost in natural accuracy. Our technique relies on the use of an Optical Processing Unit (OPU), a photonic co-processor, and a fine-tuning step performed with Direct Feedback Alignment, a synthetic gradient training scheme. We test our method on nine different models against four attacks in RobustBench, consistently improving over state-of-the-art performance. We perform an ablation study on the single components of our defense, showing that robustness arises from parameter obfuscation and the alternative training method. We also introduce phase retrieval attacks, specifically designed to increase the threat level of attackers against our own defense. We show that even with state-of-the-art phase retrieval techniques, ROPUST remains an effective defense.
\end{abstract}

\section{Introduction}
Adversarial examples \cite{Goodfellow2015ExplainingAH} threaten the safety and reliability of machine learning models deployed in the wild. Because of the sheer number of attack and defense scenarios, true real-world robustness can be difficult to evaluate~\cite{bubeck2019adversarial}. Standardized benchmarks, such as RobustBench \cite{croce2020robustbench} using AutoAttack \cite{Croce2020ReliableEO}, have helped better evaluate progress in the field. Furthermore, the development of defense-specific attacks is also crucial \cite{tramer2019adversarial}. To date, one of the most effective defense techniques remains adversarial training with Projected Gradient Descent (PGD)~\cite{madry2017towards}. Adversarial training of a model can be resource-consuming, but robust networks pre-trained with PGD are now widely available. 

This motivates the use of these pre-trained robust models as a solid foundation for developing simple and widely applicable defenses that further enhance their robustness. To this end, we introduce \textbf{ROPUST}, a drop-in replacement for the classifier of already robust models. Our defense is unique in that it leverages a photonic co-processor (the Optical Processing Unit, OPU) for physical \emph{parameter obfuscation} \cite{Cappelli2021AdversarialRB}: because the \emph{fixed} random parameters are optically implemented, they remain unknown at training and inference time. Additionally, a synthetic gradient method, Direct Feedback Alignment (DFA) \cite{Nkland2016DirectFA}, is used for fine-tuning the ROPUST classifier. 

\begin{figure}
    \centering
    \includegraphics[width=\textwidth]{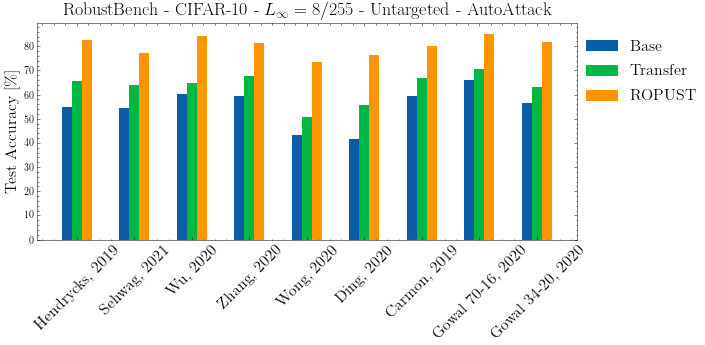}
    \caption{\textbf{ROPUST systematically improves the test accuracy of already robust models}. Transfer refers to the performance when attacks are generated on the base model and transferred to the ROPUST model. Models from the RobustBench model zoo: Hendrycks et al., 2019 \cite{Hendrycks2019UsingPC}, Sehwag et al., 2021 \cite{Sehwag2021ImprovingAR}, Wu et al., 2020 \cite{Wu2020AdversarialWP}, Zhang et al., 2020 \cite{Zhang2020GeometryawareIA}, Wong et al., 2020 \cite{Wong2020FastIB}, Ding et al., 2020 \cite{Ding2020MaxMarginA}, Carmon et al., 2019 \cite{Carmon2019UnlabeledDI}, Gowal et al., 2020 \cite{Gowal2020UncoveringTL}.}
    \label{fig:robustbench}
\end{figure}

We evaluate extensively our method against AutoAttack on nine different models in RobustBench, and consistently improve robust accuracies over the state-of-the-art (Section \ref{sec:robustbench} and Fig. \ref{fig:robustbench}). We perform an ablation study, in Section \ref{sec:ablation}, and find that the robustness of our defense against white-box attacks comes from both \emph{parameter obfuscation} and DFA. Surprisingly, we also discover that simply retraining the classifier of a robust model on natural data increases its robustness to square attacks, a phenomenon that warrants further study. Finally, in Section \ref{sec:phase_retrieval}, we develop a \emph{phase retrieval} attack targeting specifically the parameter obfuscation of our defense, and show that even against state-of-the art phase retrieval techniques, ROPUST achieves fair robustness.

\subsection{Related work}
\paragraph{Attacks.} Adversarial attacks have been framed in a variety of settings: white-box, where the attacker is assumed to have unlimited access to the model, including its parameters (e.g. FGSM \cite{Goodfellow2015ExplainingAH}, PGD \cite{madry2017towards, kurakin2016adversarial}, Carlini \& Wagner \cite{carlini2017towards}); black-box, assuming only limited access to the network for the attacker, such as the label or logits for a given input, with methods attempting to estimate the gradients \cite{chen2017zoo,ilyas2018black,ilyas2018prior}, or more recently derived from genetic algorithms \cite{andriushchenko2019square,meunier2019yet} and combinatorial optimization \cite{moon19aparsimonous}; transfer attacks, where an attack is crafted on a similar model that is accessible to the attacker, and then applied to the target network \cite{papernot2016transferability}. Automated schemes, such as AutoAttack \cite{Croce2020ReliableEO}, have been proposed to  autonomously select which attack to perform against a given network, and to automatically tune its hyperparameters. 

\paragraph{Defenses.} Adversarial training adds adversarial robustness as an explicit training objective \cite{Goodfellow2015ExplainingAH,madry2017towards}, by incorporating adversarial examples during the training. This has been, and still is, one of the most effective defense against attacks. Repository of pre-trained robust models have been compiled, such as the RobustBench Model Zoo\footnote{Accessible at: \url{https://github.com/RobustBench/robustbench}.}. Conversely, theoretically grounded defenses have been proposed \citep{lecuyer2018certified,KolterRandomizedSmoothing,araujo2020,pinot2019theoretical,wong2018scaling,wong2018provable}, but these fail to match the clean accuracy of state-of-the-art networks, making robustness a trade-off with performance. Many empirical defenses have been criticized for providing a false sense of security \cite{athalye2018obfuscated,tramer2019adversarial}, by not evaluating on attacks adapted to the defense.  

\paragraph{Obfuscation.} Gradient obfuscation, through the use of a non-differentiable activation function, has been proposed as a way to protect against white-box attacks \cite{papernot2017practical}. However, gradient obfuscation can be easily bypassed by Backward Pass Differentiable Approximation (BPDA) \cite{athalye2018obfuscated}, where the defense is replaced by an approximated and differentiable version. \emph{Parameter obfuscation} has been proposed with dedicated photonic co-processor \cite{Cappelli2021AdversarialRB}, enforced by the physical properties of said co-processor. However, by itself, this kind of defense falls short of adversarial training. 

\paragraph{Fine-tuning and analog computing.} Previous work introduced \textit{adversarial fine-tuning} \cite{Jeddi2020ASF}: fine-tuning a non-robust model with an adversarial objective. In this work instead we fine-tune a robust model without adversarial training. Additionally, it was shown that robustness improves transfer performance \cite{Salman2020DoAR} and that robustness transfers across datasets \cite{Shafahi2020AdversariallyRT}. The advantage of non-ideal analog computations in terms of robustness has been investigated in the context of NVM crossbars \cite{roy2020robustness}, while we here focus on a photonic technology, readily available to perform computations at scale.

\subsection{Motivations and contributions}

We propose to simplify and extend the applicability of photonic-based parameter obfuscation defenses. Our defense, ROPUST, is a universally and easily applicable drop-in replacement for classifiers of already robust models. In contrast with existing parameter-obfuscation methods, it leverages pre-trained robust models, and achieves state-of-the-art performance. 

\paragraph{Beyond silicon and beyond backpropagation.} We leverage photonic hardware and alternative training methods to achieve adversarial robustness. The use of dedicated hardware to perform the random projection physically guarantees \emph{parameter obfuscation}. Direct Feedback Alignment enables us to train and/or fine-tune the model despite non-differentiable analog hardware being used in the forward pass. In our ablation study, we find that both these components contribute to adversarial robustness, providing a holistic defense.

\paragraph{Simple, universal, and state-of-the-art.} ROPUST can be dropped-in to supplement any robust pre-trained model, replacing its classifier. Fine-tuning the ROPUST classifier is fast and does not require additional changes to the model architecture. This enables any existing architecture and adversarial countermeasure to leverage ROPUST to gain additional robustness, at limited cost.  We evaluate on RobustBench, across 9 pre-trained models, against AutoAttack sampling from a pool of 4 attacks. We achieve state-of-the-art performance on the leaderboard, and, in light of our results, we suggest the extension of RobustBench to include obfuscation-based methods. 

\paragraph{The Square attack mystery.} Performing an ablation study on Square attack~\cite{andriushchenko2019square}, we find that simply retraining from scratch the classifier of a robust model on natural data increases its robustness against it. This phenomenon remains unexplained and occurs even when the original fully connected classification layer is retrained, without using our ROPUST module.

\paragraph{Phase retrieval attacks.} Drawing inspiration from the field of phase retrieval, we introduce a new kind of attack against defenses relying on parameter obfuscation, \emph{phase retrieval attacks}. These attacks assume the attacker leverage phase retrieval techniques to retrieve the obfuscated parameters in full, and we show that ROPUST remains robust even against state-of-the-art retrieval methods.

\section{Methods}\label{sec:methods}

\subsection{Automated adversarial attacks}
We evaluate our model against the four attacks implemented in RobustBench: APGD-CE and APGD-T \cite{Croce2020ReliableEO}, Square attack \cite{andriushchenko2019square}, and Fast Adaptive Boundary (FAB) attack \cite{Croce2020MinimallyDA}. APGD-CE is a standard PGD where the step size is tuned using the loss trend information, squeezing the best performance out of a limited iterations budget. APGD-T, on top of the step size schedule, substitutes the cross-entropy loss with the Difference of Logits Ratio (DLR) loss, reducing the risk of vanishing gradients. Square attack is based on a random search. Random updates $\delta$ are sampled from an attack-norm dependent distribution at each iteration: if they improve the objective function they are kept, otherwise they are discarded. FAB attack aims at finding adversarial samples with minimal distortion with respect to the attack point. With respect to PGD, it does not need to be restarted and it achieves fast good quality results. In RobustBench, using AutoAttack, given a batch of samples, these are first attacked with APGD-CE. Then, the samples that were successfully attacked are discarded, and the remaining ones are attacked with APGD-T. This procedure continues with Square and FAB attack.

\subsection{Our defense}\label{sec:defense}
\begin{figure}
    \centering
    \includegraphics[width=\textwidth]{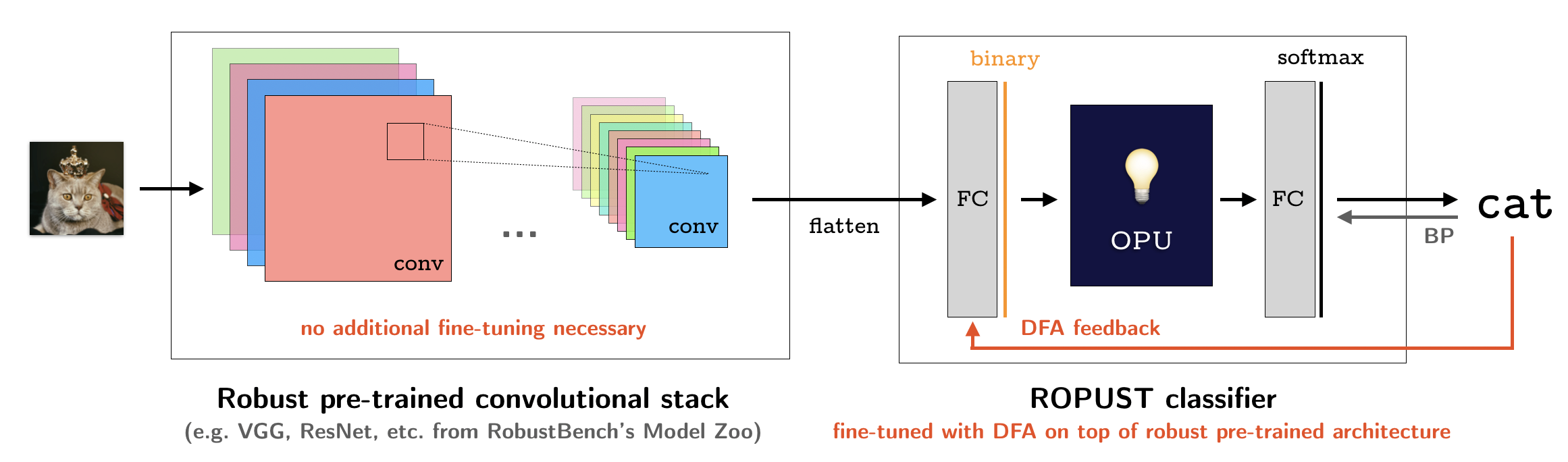}
    \caption{\textbf{ROPUST replaces the classifier of already robust models, enhancing their adversarial robustness.} Only the ROPUST classifier needs fine-tuning; the convolutional stack is frozen. Convolutional features first go through a fully-connected layer, before binarization for use in the Optical Processing Unit (OPU). The OPU performs a non-linear random projection, with \emph{fixed unknown parameters}. A fully-connected layer is then used to obtain a prediction from the output of the OPU. Direct Feedback Alignment is used to train the layer underneath the OPU.}
    \label{fig:ropust}
\end{figure}
\paragraph{Optical Processing Units.} Optical Processing Units (OPU)\footnote{Accessible through LightOn Cloud: \url{https://cloud.lighton.ai}.} are photonic co-processors dedicated to efficient large-scale random projections. Assuming an input vector $\mathbf{x}$, the OPU computes the following operation using light scattering through a diffusive medium:
\begin{equation}\label{eq:opu}
    \mathbf{y} = \lvert \mathbf{Ux}\rvert^2
\end{equation}

With $\textbf{U}$ a \emph{fixed} complex Gaussian random matrix of size up to $10^6\times 10^6$, which entries are not readily known. In the following, we sometimes refer to $\mathbf{U}$ as the \emph{transmission matrix} (TM). The input $\textbf{x}$ is binary (1 bit -- 0/1) and the output $\textbf{y}$ is quantized in 8-bit. While it is possible to simulate an OPU and implement ROPUST on GPU, this comes with two significant drawbacks: (1) part of our defense relies on $\mathbf{U}$ being obfuscated to the attacker, which is not possible to guarantee on a GPU; (2) at large sizes, storing $\mathbf{U}$ in GPU memory is costly \cite{ohana2020kernel}. 

Because $\mathbf{U}$ is physically implemented through the diffusive medium, the random matrix will remain unknown even if the host system is compromised. Assuming unfettered access to the OPU, an attacker has to perform \emph{phase retrieval} to retrieve the coefficients of $\mathbf{U}$. As only the non-linear intensity $\lvert\mathbf{Ux}\rvert^2$ can be measured and not $\mathbf{Ux}$ directly, this phase retrieval step is computationally costly. This problem is well studied, and state-of-the-art methods have $O(MN\log N)$ time complexity \cite{Gupta2020FastOS}, and do not result in a perfect retrieval. We develop an attack scenario based on this method in Section \ref{sec:phase_retrieval}.

\paragraph{Direct Feedback Alignment.} Because the fixed random parameters implemented by the OPU are unknown, it is impossible to backpropagate through it. We bypass this limitation by training layers upstream of the OPU using Direct Feedback Alignment (DFA) \cite{Nkland2016DirectFA}. DFA is an alternative to backpropagation, capable of scaling to modern deep learning tasks and architectures \cite{Launay2020DirectFA}, relying on a random projection of the error as the teaching signal.

In a fully connected network, at layer $i$ out of $N$, neglecting biases, with $\mathbf{W}_i$ its weight matrix, $f_i$ its non-linearity, and $\mathbf{h}_i$ its activations, the forward pass can be written as $\mathbf{a}_i = \mathbf{W}_i \mathbf{h}_{i - 1}, \mathbf{h}_i = f_i(\mathbf{a}_i)$.
$\mathbf{h}_0 = X$ is the input data, and $\mathbf{h}_N = f(\mathbf{a}_N) = \mathbf{\hat{y}}$ are the predictions.  A task-specific cost function $\mathcal{L}(\mathbf{\hat{y}}, \mathbf{y})$ is computed to quantify the quality of the predictions with respect to the targets $\mathbf{y}$. The weight updates are obtained through the chain-rule of derivatives: 
\begin{equation}
    \delta \mathbf{W}_i = - \frac{\partial \mathcal{L}}{\partial \mathbf{W}_i} = - [(\mathbf{W}_{i+1}^{T} \delta \mathbf{a}_{i+1})\odot f'_i(\mathbf{a}_i)] \mathbf{h}_{i-1}^{T}, \delta \mathbf{a}_i = \frac{\partial \mathcal{L}}{\partial \mathbf{a}_i}
\end{equation}

where $\odot$ is the Hadamard product. With DFA, the gradient signal $\mathbf{W}_{i+1}^{T}\delta \mathbf{a}_{i + 1}$ of the (i+1)-th layer is replaced with a random projection of the gradient of the loss at the top layer $\delta \mathbf{a}_y$--which is the error $\mathbf{e} = \mathbf{\hat{y}} - \mathbf{y}$ for commonly used losses, such as cross-entropy or mean squared error: 
\begin{equation}
    \delta \mathbf{W}_i = - [(\mathbf{B}_i\delta \mathbf{a}_y) \odot f'_i(\mathbf{a}_i)] \mathbf{h}_{i-1}^T, \delta \mathbf{a}_y = \frac{\mathbf{\partial \mathcal{L}}}{\partial \mathbf{a}_y}
\end{equation}

Learning with DFA is enabled by an alignment process, wherein the forward weights learn a configuration enabling DFA to approximate BP updates \cite{refinetti2020dynamics}.

\paragraph{ROPUST} 
To enhance the adversarial robustness of pretrained robust models, we propose to replace their classifier with the ROPUST module (Fig. \ref{fig:ropust}). We use robust models from the RobustBench model zoo, extracting and freezing their convolutional stack. The robust convolutional features go through a fully connected layer and a binarization step (a sign function), preparing them for the OPU. The OPU then performs a non-linear random projection,  with fixed unknown parameters. Lastly, the predictions are obtained through a final fully-connected layer. While the convolutional layers are frozen, we train the ROPUST module on natural data using DFA to bypass the non-differentiable photonic hardware. 

\paragraph{Attacking ROPUST.} While we could use DFA to attack ROPUST, previous work has shown that methods devoid of weight transport are not effective in generating compelling adversarial examples~\cite{Akrout2019OnTA}. Therefore, we instead use backward pass differentiable approximation (BPDA) when attacking our defense. For BPDA, we need to find a good differentiable relaxation to non-differentiable layers. For the binarization function, we simply use the derivative of $\tanh$ in the backward pass, while we approximate the transpose of the obfuscated parameters with a different fixed random matrix drawn at initialization of the module. More specifically, if we consider the expression for the forward pass of the ROPUST module:
\begin{equation}
    \mathbf{y} = \mbox{softmax}(\mathbf{W}_3\lvert\mathbf{U} \mbox{sign}(\mathbf{W}_1 \mathbf{x})\rvert^2)
\end{equation}
In the backward we substitute $\mathbf{U}^T$ (that we do not have access to) with a different fixed random matrix $\mathbf{R}$, in a setup similar to Feedback Alignment \cite{Lillicrap2014RandomFW}. We also relax the sign function derivative to the derivative of $\tanh$.

We present empirical results on RobustBench in the following Section \ref{sec:robustbench}. We then ablate the components of our defense in \ref{sec:ablation}, demonstrating its holistic nature, and we finally create a phase retrieval attack to challenge parameter obfuscation in Section \ref{sec:phase_retrieval}.

\section{Evaluating ROPUST on RobustBench}\label{sec:robustbench}
\begin{figure}
    \centering
    \includegraphics[width=\textwidth]{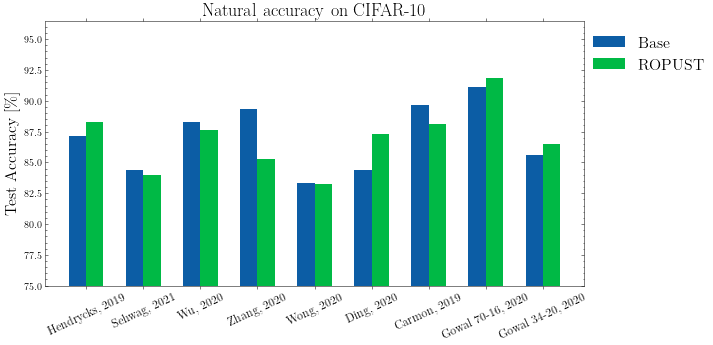}
    \caption{\textbf{Our ROPUST defense comes at no cost in natural accuracy.} In some cases, natural accuracy is even improved. The model from Zhang, 2020 \cite{Zhang2020GeometryawareIA} is an isolated exception. Models from the RobustBench model zoo: Hendrycks et al., 2019 \cite{Hendrycks2019UsingPC}, Sehwag et al., 2021 \cite{Sehwag2021ImprovingAR}, Wu et al., 2020 \cite{Wu2020AdversarialWP}, Zhang et al., 2020 \cite{Zhang2020GeometryawareIA}, Wong et al., 2020 \cite{Wong2020FastIB}, Ding et al., 2020 \cite{Ding2020MaxMarginA}, Carmon et al., 2019 \cite{Carmon2019UnlabeledDI}, Gowal et al., 2020 \cite{Gowal2020UncoveringTL}.}
    \label{fig:natacc}
\end{figure}
All of the attacks are performed on CIFAR-10 \cite{Krizhevsky2009LearningML}, using a differentiable backward pass approximation \cite{athalye2018obfuscated} as explained in Section \ref{sec:defense}. For our experiments, we use OPU input size $512$ and output size $8000$. We use the Adam optimizer \cite{kingma2014adam}, with learning rate $0.001$, for $10$ epochs. The process typically takes as little as 10 minutes on a single NVIDIA V100 GPU. 

We show our results on nine different models in RobustBench in Fig. \ref{fig:robustbench}. The performance of the original pretrained models from the RobustBench leaderboard is reported as \textit{Base}. \textit{ROPUST} represents the same models equipped with our defense. Finally, \textit{Transfer} indicates the performance of attacks created on the original model and transferred to fool the ROPUST defense. For all models considered, ROPUST improves the robustness significantly, even under transfer. 

For transfer, we also tested crafting the attacks on the \textit{Base} model while using the loss of the ROPUST model for the learning rate schedule of APGD. We also tried to use the predictions of ROPUST, instead of the base model, to \textit{remove} the samples that were successfully attacked from the next stage of the ensemble; however, these modifications did not improve transfer performance.

Finally, we remark that the robustness increase typically comes at no cost in natural accuracy; we show the accuracy on natural data of the \textit{Base} and the \textit{ROPUST} models in Fig. \ref{fig:natacc}.

\section{Understanding ROPUST: an ablation study}\label{sec:ablation}
\begin{figure}
    \centering
    \includegraphics[width=0.9\textwidth]{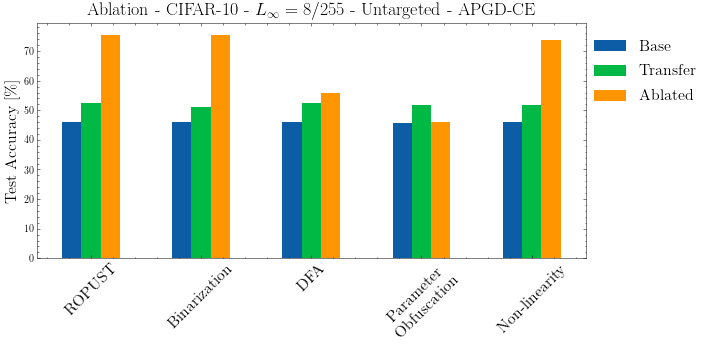}
    \caption{\textbf{Removing either parameter obfuscation or DFA from our defense causes a large drop in accuracy.} This confirms the intuition that robustness is given by the inability to efficiently generate attacks in a white-box settings when the parameters are obfuscated, and that DFA is capable of generating partially robust features. We note that even though the non-linearity $|.|^2$ does not contribute to robustness, it is key to obfuscation, preventing trivial retrieval. Transfer performance does not change much when removing components of the defense. While the \texttt{Base} model is not ablated, we leave its performance as a term of comparison.}
    \label{fig:ablation}
\end{figure}
\begin{figure}
    \centering
    \includegraphics[width=0.9\textwidth]{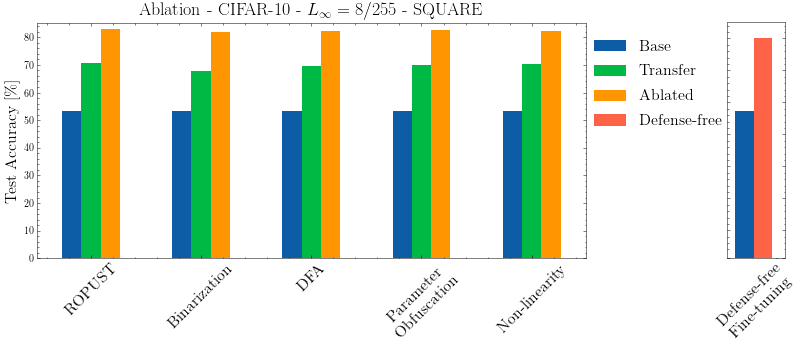}
    \caption{\textbf{Square attack can be evaded by simply retraining on natural data the classifier of a robust model.} We confirm the same result when retraining the standard fully connected classification layer in the pretrained models in place of the ROPUST module (\textit{Defense-free} result in the chart on the right). While the \texttt{Base} model is not ablated, we leave its performance as a term of comparison.}
    \label{fig:ablation_square}
\end{figure}
We use the model from \cite{Wong2020FastIB} available in the RobustBench model zoo to perform our ablation studies. It consists in a PreAct ResNet-18 \cite{He2016IdentityMI}, pretrained with a \textit{"revisited"} FGSM of increased effectiveness.
\paragraph{Holistic defense.}
We conduct an ablation study by removing a single component of our defense at a time in simulation: binarization, DFA, parameter obfuscation, and non-linearity $|.|^2$ of the random projection. To remove DFA, we also remove the binarization step and train the ROPUST module with backpropagation, since we have access to the transpose of the transmission matrix in the simulated setting of the ablation study. We show the results in Fig. \ref{fig:ablation}: we see that removing the non-linearity $|.|^2$ and the binarization does not have an effect, with the robustness given by \textit{parameter obfuscation} and DFA, as expected. However, note that $|.|^2$ is central to preventing trivial phase retrieval, and is hence a key component of our defense.  
\paragraph{Robustness to Square attack} 
While the ablation study on the APGD attack is able to pinpoint the exact sources of robustness for a white-box attack, the same study on the black-box Square attack has surprising results. Indeed, as shown in Fig. \ref{fig:ablation_square}, no element of the ROPUST mechanism can be linked to robustness against Square attack. Interestingly, we found an identical behaviour when retraining the standard fully connected classification layer from scratch on natural (non perturbed) data, shown in the same Fig. \ref{fig:ablation_square} under the \textit{Defense-free} label.

\section{Phase retrieval attack}\label{sec:phase_retrieval}

Our defense leverages parameter obfuscation to achieve robustness. Yet, however demanding, it is still technically possible to recover the parameters through phase retrieval schemes \cite{Gupta2019DontTI,Gupta2020FastOS}. To provide a thorough and fair evaluation of our attack, we study in this section \emph{phase retrieval} attacks. We first consider an idealized setting, and then confront this setting with a real-world phase retrieval algorithm from~\cite{Gupta2020FastOS}.
\begin{figure}
    \centering
    \includegraphics[width=0.9\textwidth]{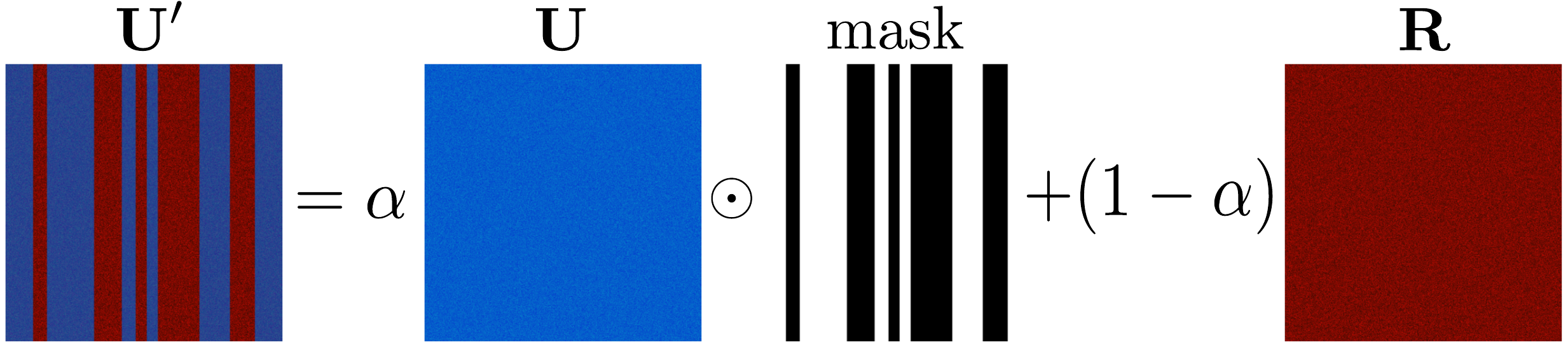}
    \caption{\textbf{Simplified modelling of phase retrieval.} The retrieved matrix $\mathbf{U'}$ is modeled as the linear interpolation between the real transmission matrix $\mathbf{U}$ and a random matrix $\mathbf{R}$, only for some columns selected by a mask. Varying the value of $\alpha$ and the percentage of masked columns allows to modulate the knowledge of the attacker without running resource-hungry phase retrieval algorithms.}
    \label{fig:retrieval-model}
\end{figure}
\paragraph{Ideal retrieval model.} We build an idealized phase retrieval attack, where the attacker knows a certain fraction of columns, up to a certain precision, schematized in Figure \ref{fig:retrieval-model}. To smoothly vary the precision, we model the retrieved matrix $\mathbf{U'}$ as a linear interpolation of the real transmission matrix $\mathbf{U}$ and a completely different random matrix $\mathbf{R}$:
\begin{equation}
    \mathbf{U'} = \alpha\mathbf{U} + (1-\alpha)\mathbf{R}
\end{equation}
In real phase retrieval, this model is valid for a certain fraction of columns of the transmission matrix, and the remaining ones are modeled as independent random vectors. We can model this with a Boolean mask matrix $\mathbf{M}$, so our retrieval model in the end is:
\begin{equation}
    \mathbf{U'} = \alpha\mathbf{U}\odot\mathbf{M} + (1-\alpha)\mathbf{R}
\end{equation}
\begin{figure}
    \centering
    \includegraphics[width=0.5\textwidth]{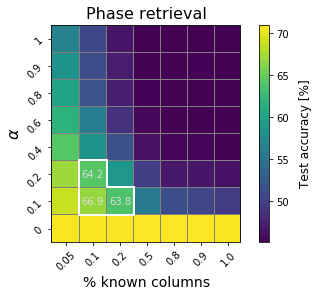}
    \caption{\textbf{Performance of an APGD-CE attack with a retrieved matrix in place of the, otherwise unknown, transpose of the transmission matrix.} As expected, a better knowledge of the transmission matrix, i.e. higher alpha and/or higher percentage of known columns correlates with the success of the attack, with a sharp phase transition. At first glance, it may seem that even a coarse-grained knowledge of the TM can help the attacker. However, optical phase retrieval works on the output correlation only: accordingly, we find that even state-of-the-art phase retrieval methods operates only in the white contoured region, where the robustness is still greater than the \textit{Base} models. We highlighted the accuracies achieved under attack in this region in the heat-map.}
    \label{fig:phase_retrieval}
\end{figure}
In this setting, we vary the knowledge of the attacker from the minimum to the maximum by varying $\alpha$ and the percentage of retrieved columns, and we show how the performance of our defense changes in Fig. \ref{fig:phase_retrieval}. In this simplified model only a crude knowledge of the parameters seems sufficient, given the sharp phase transition. We now need to chart where state-of-the-art retrieval methods are on this graph to estimate their ability to break our defense.

\paragraph{Real-world retrieval performance.} State-of-the-art phase retrieval methods seek to maximize output correlation, i.e. the correlation on $\mathbf{y}$ in Eq. \ref{eq:opu}, in place of the correlation with respect to the parameters of the transmission matrix, i.e. $\mathbf{U}$ in Eq. \ref{eq:opu}. This leads to a retrieved matrix that may well approximate the OPU outputs, but not the actual transmission matrix it implements. We find this is a significant limitation for attackers. In Fig. \ref{fig:phase_retrieval}, following numerical experiments, we highlight with a white contour the operating region of a state-of-the-art phase retrieval algorithm \cite{Gupta2020FastOS}, showing that it can manage to only partially reduce the robustness of ROPUST. 
\section{Conclusion}
We introduced ROPUST, a drop-in module to enhance the adversarial robustness of pretrained already robust models. Our technique relies on parameter obfuscation guaranteed by a photonic co-processor, and a synthetic gradient method: it is simple, fast and widely applicable.

We thoroughly evaluated our defense on nine different models in the standardized RobustBench benchmark, reaching state-of-the-art performance. In light of these results, we encourage to extend RobustBench to include parameter obfuscation methods. 

We performed an ablation study in the white-box setting, confirming our intuition and the results from \cite{Cappelli2021AdversarialRB}: the robustness comes from the parameter obfuscation and from the hybrid synthetic gradient method. The non-linearity $|.|^2$ on the random projection, while not contributing to robustness on its own, is key to prevent trivial deobfuscation by hardening ROPUST against phase retrieval.  A similar study in the black-box setting was inconclusive. However it shed light on a phenomenon of increased robustness against Square attack when retraining from scratch the classifier of robust architectures on natural data. This phenomenon appears to be universal, i.e. independent of the structure of the classification module being fine-tuned, warranting further study. 

Finally, we developed a new kind of attacks, \textit{phase retrieval attacks}, specifically suited to parameter obfuscation defense such as ours, and we tested their effectiveness. We found that the typical precision regime of even state-of-the-art phase retrieval methods is not enough to completely break ROPUST.

Future work could investigate how the robustness varies with the input and output size of the ROPUST module, and if there are different parameter obfuscation trade-offs when such dimensions change. The combination of ROPUST with other defense techniques, such as adversarial label-smoothing~\cite{Goibert2019AdversarialRV}, could also be of interest to further increase robustness. By combining beyond silicon hardware and beyond backpropagation training methods, our work highlights the importance of considering solutions outside of the hardware lottery \cite{Hooker2020TheHL}. 

\paragraph{Broader impact.} Adversarial attacks have been identified as a significant threat to applications of machine learning in-the-wild. Developing simple and accessible ways to make neural networks more robust is key to mitigating some of the risks and making machine learning applications safer. In particular, more robust models would enable a wider range of business applications, especially in safety-critical sectors. 

We do not foresee negative societal impacts from our work, beyond the risk of our defense being broken by future developments of research in adversarial attacks.

A limit of our work is that we prove increased robustness only empirically and not theoretically. However, we note that theoretically grounded defense methods typically fall short of other techniques more used in practice. We also rely on photonic hardware, that is however accessible by anyone similarly to GPUs or TPUs on commercial cloud providers.

We performed all of our experiments on single-GPU nodes with NVIDIA V100, and an OPU, on a cloud provider. We estimate a total of $\sim 500$ GPU hours was spent.

\begin{ack}
The authors thank the LightOn Team for their useful feedback and support. Ruben Ohana acknowledges funding from the R\'egion Ile-de-France.
\end{ack}

\bibliographystyle{unsrt}
\bibliography{references}

\begin{thebibliography}{10}

\bibitem{Goodfellow2015ExplainingAH}
I.~Goodfellow, Jonathon Shlens, and Christian Szegedy.
\newblock Explaining and harnessing adversarial examples.
\newblock {\em CoRR}, abs/1412.6572, 2015.

\bibitem{bubeck2019adversarial}
S{\'e}bastien Bubeck, Yin~Tat Lee, Eric Price, and Ilya Razenshteyn.
\newblock Adversarial examples from computational constraints.
\newblock In {\em International Conference on Machine Learning}, pages
  831--840. PMLR, 2019.

\bibitem{croce2020robustbench}
Francesco Croce, Maksym Andriushchenko, Vikash Sehwag, Nicolas Flammarion, Mung
  Chiang, Prateek Mittal, and Matthias Hein.
\newblock Robustbench: a standardized adversarial robustness benchmark.
\newblock {\em arXiv preprint arXiv:2010.09670}, 2020.

\bibitem{Croce2020ReliableEO}
Francesco Croce and Matthias Hein.
\newblock Reliable evaluation of adversarial robustness with an ensemble of
  diverse parameter-free attacks.
\newblock In {\em ICML}, 2020.

\bibitem{tramer2019adversarial}
Florian Tram{\`e}r and Dan Boneh.
\newblock Adversarial training and robustness for multiple perturbations.
\newblock In {\em Advances in Neural Information Processing Systems}, pages
  5866--5876, 2019.

\bibitem{madry2017towards}
Aleksander Madry, Aleksandar Makelov, Ludwig Schmidt, Dimitris Tsipras, and
  Adrian Vladu.
\newblock Towards deep learning models resistant to adversarial attacks.
\newblock In {\em International Conference on Learning Representations}, 2018.

\bibitem{Cappelli2021AdversarialRB}
A.~Cappelli, Ruben Ohana, Julien Launay, Laurent Meunier, Iacopo Poli, and
  F.~Krzakala.
\newblock Adversarial robustness by design through analog computing and
  synthetic gradients.
\newblock {\em ArXiv}, abs/2101.02115, 2021.

\bibitem{Nkland2016DirectFA}
Arild N{\o}kland.
\newblock Direct feedback alignment provides learning in deep neural networks.
\newblock In {\em NIPS}, 2016.

\bibitem{Hendrycks2019UsingPC}
Dan Hendrycks, Kimin Lee, and Mantas Mazeika.
\newblock Using pre-training can improve model robustness and uncertainty.
\newblock In {\em ICML}, 2019.

\bibitem{Sehwag2021ImprovingAR}
V.~Sehwag, Saeed Mahloujifar, Tinashe Handina, Sihui Dai, Chong Xiang,
  M.~Chiang, and Prateek Mittal.
\newblock Improving adversarial robustness using proxy distributions.
\newblock {\em ArXiv}, abs/2104.09425, 2021.

\bibitem{Wu2020AdversarialWP}
Dongxian Wu, Shutao Xia, and Yisen Wang.
\newblock Adversarial weight perturbation helps robust generalization.
\newblock {\em arXiv: Learning}, 2020.

\bibitem{Zhang2020GeometryawareIA}
Jingfeng Zhang, Jianing Zhu, Gang Niu, B.~Han, M.~Sugiyama, and M.~Kankanhalli.
\newblock Geometry-aware instance-reweighted adversarial training.
\newblock {\em ArXiv}, abs/2010.01736, 2020.

\bibitem{Wong2020FastIB}
Eric Wong, Leslie Rice, and J.~Z. Kolter.
\newblock Fast is better than free: Revisiting adversarial training.
\newblock {\em ArXiv}, abs/2001.03994, 2020.

\bibitem{Ding2020MaxMarginA}
Gavin~Weiguang Ding, Yash Sharma, Kry Yik-Chau Lui, and Ruitong Huang.
\newblock Max-margin adversarial (mma) training: Direct input space margin
  maximization through adversarial training.
\newblock {\em ArXiv}, abs/1812.02637, 2020.

\bibitem{Carmon2019UnlabeledDI}
Y.~Carmon, Aditi Raghunathan, Ludwig Schmidt, Percy Liang, and John~C. Duchi.
\newblock Unlabeled data improves adversarial robustness.
\newblock In {\em NeurIPS}, 2019.

\bibitem{Gowal2020UncoveringTL}
Sven Gowal, Chongli Qin, Jonathan Uesato, Timothy~A. Mann, and P.~Kohli.
\newblock Uncovering the limits of adversarial training against norm-bounded
  adversarial examples.
\newblock {\em ArXiv}, abs/2010.03593, 2020.

\bibitem{kurakin2016adversarial}
Alexey Kurakin, Ian Goodfellow, and Samy Bengio.
\newblock Adversarial examples in the physical world.
\newblock {\em arXiv preprint arXiv:1607.02533}, 2016.

\bibitem{carlini2017towards}
Nicholas Carlini and David Wagner.
\newblock Towards evaluating the robustness of neural networks.
\newblock In {\em 2017 IEEE Symposium on Security and Privacy (SP)}, pages
  39--57. IEEE, 2017.

\bibitem{chen2017zoo}
Pin-Yu Chen, Huan Zhang, Yash Sharma, Jinfeng Yi, and Cho-Jui Hsieh.
\newblock Zoo: Zeroth order optimization based black-box attacks to deep neural
  networks without training substitute models.
\newblock In {\em Proceedings of the 10th ACM Workshop on Artificial
  Intelligence and Security}, pages 15--26. ACM, 2017.

\bibitem{ilyas2018black}
Andrew Ilyas, Logan Engstrom, Anish Athalye, and Jessy Lin.
\newblock Black-box adversarial attacks with limited queries and information.
\newblock {\em arXiv preprint arXiv:1804.08598}, 2018.

\bibitem{ilyas2018prior}
Andrew Ilyas, Logan Engstrom, and Aleksander Madry.
\newblock Prior convictions: Black-box adversarial attacks with bandits and
  priors.
\newblock {\em arXiv preprint arXiv:1807.07978}, 2018.

\bibitem{andriushchenko2019square}
Maksym Andriushchenko, Francesco Croce, Nicolas Flammarion, and Matthias Hein.
\newblock Square attack: a query-efficient black-box adversarial attack via
  random search.
\newblock {\em arXiv preprint arXiv:1912.00049}, 2019.

\bibitem{meunier2019yet}
Laurent Meunier, Jamal Atif, and Olivier Teytaud.
\newblock Yet another but more efficient black-box adversarial attack: tiling
  and evolution strategies.
\newblock {\em arXiv preprint arXiv:1910.02244}, 2019.

\bibitem{moon19aparsimonous}
Seungyong Moon, Gaon An, and Hyun~Oh Song.
\newblock Parsimonious black-box adversarial attacks via efficient
  combinatorial optimization.
\newblock In Kamalika Chaudhuri and Ruslan Salakhutdinov, editors, {\em
  Proceedings of the 36th International Conference on Machine Learning},
  volume~97 of {\em Proceedings of Machine Learning Research}, pages
  4636--4645, Long Beach, California, USA, 09--15 Jun 2019. PMLR.

\bibitem{papernot2016transferability}
Nicolas Papernot, Patrick McDaniel, and Ian Goodfellow.
\newblock Transferability in machine learning: from phenomena to black-box
  attacks using adversarial samples.
\newblock {\em arXiv preprint arXiv:1605.07277}, 2016.

\bibitem{lecuyer2018certified}
M.~Lecuyer, V.~Atlidakis, R.~Geambasu, D.~Hsu, and S.~Jana.
\newblock Certified robustness to adversarial examples with differential
  privacy.
\newblock In {\em 2019 IEEE Symposium on Security and Privacy (SP)}, pages
  727--743, 2018.

\bibitem{KolterRandomizedSmoothing}
Jeremy~M. Cohen, Elan Rosenfeld, and J.~Zico Kolter.
\newblock Certified adversarial robustness via randomized smoothing.
\newblock {\em arXiv preprint arXiv:1902.02918}, abs/1902.02918, 2019.

\bibitem{araujo2020}
Rafael~Pinot Alexandre~Araujo, Laurent~Meunier and Benjamin Negrevergne.
\newblock Advocating for multiple defense strategies against adversarial
  examples.
\newblock {\em Workshop on Machine Learning for CyberSecurity
  (MLCS@ECML-PKDD)}, 2020.

\bibitem{pinot2019theoretical}
Rafael Pinot, Laurent Meunier, Alexandre Araujo, Hisashi Kashima, Florian Yger,
  C{\'e}dric Gouy-Pailler, and Jamal Atif.
\newblock Theoretical evidence for adversarial robustness through
  randomization: the case of the exponential family.
\newblock {\em arXiv preprint arXiv:1902.01148}, 2019.

\bibitem{wong2018scaling}
Eric Wong, Frank Schmidt, Jan~Hendrik Metzen, and J~Zico Kolter.
\newblock Scaling provable adversarial defenses.
\newblock In {\em Advances in Neural Information Processing Systems}, pages
  8400--8409, 2018.

\bibitem{wong2018provable}
Eric Wong and Zico Kolter.
\newblock Provable defenses against adversarial examples via the convex outer
  adversarial polytope.
\newblock In {\em International Conference on Machine Learning}, pages
  5286--5295, 2018.

\bibitem{athalye2018obfuscated}
Anish Athalye, Nicholas Carlini, and David Wagner.
\newblock Obfuscated gradients give a false sense of security: Circumventing
  defenses to adversarial examples.
\newblock In Jennifer Dy and Andreas Krause, editors, {\em Proceedings of the
  35th International Conference on Machine Learning}, volume~80 of {\em
  Proceedings of Machine Learning Research}, pages 274--283, Stockholmsmässan,
  Stockholm Sweden, 10--15 Jul 2018. PMLR.

\bibitem{papernot2017practical}
Nicolas Papernot, Patrick McDaniel, Ian Goodfellow, Somesh Jha, Z~Berkay Celik,
  and Ananthram Swami.
\newblock Practical black-box attacks against machine learning.
\newblock In {\em Proceedings of the 2017 ACM on Asia conference on computer
  and communications security}, pages 506--519, 2017.

\bibitem{Jeddi2020ASF}
Ahmadreza Jeddi, M.~Shafiee, and A.~Wong.
\newblock A simple fine-tuning is all you need: Towards robust deep learning
  via adversarial fine-tuning.
\newblock {\em ArXiv}, abs/2012.13628, 2020.

\bibitem{Salman2020DoAR}
Hadi Salman, Andrew Ilyas, L.~Engstrom, Ashish Kapoor, and A.~Madry.
\newblock Do adversarially robust imagenet models transfer better?
\newblock {\em ArXiv}, abs/2007.08489, 2020.

\bibitem{Shafahi2020AdversariallyRT}
A.~Shafahi, Parsa Saadatpanah, C.~Zhu, Amin Ghiasi, C.~Studer, D.~Jacobs, and
  T.~Goldstein.
\newblock Adversarially robust transfer learning.
\newblock {\em ArXiv}, abs/1905.08232, 2020.

\bibitem{roy2020robustness}
Deboleena Roy, Indranil Chakraborty, Timur Ibrayev, and Kaushik Roy.
\newblock Robustness hidden in plain sight: Can analog computing defend against
  adversarial attacks?
\newblock {\em arXiv preprint arXiv:2008.12016}, 2020.

\bibitem{Croce2020MinimallyDA}
Francesco Croce and Matthias Hein.
\newblock Minimally distorted adversarial examples with a fast adaptive
  boundary attack.
\newblock In {\em ICML}, 2020.

\bibitem{ohana2020kernel}
Ruben Ohana, Jonas Wacker, Jonathan Dong, S{\'e}bastien Marmin, Florent
  Krzakala, Maurizio Filippone, and Laurent Daudet.
\newblock Kernel computations from large-scale random features obtained by
  optical processing units.
\newblock In {\em ICASSP 2020-2020 IEEE International Conference on Acoustics,
  Speech and Signal Processing (ICASSP)}, pages 9294--9298. IEEE, 2020.

\bibitem{Gupta2020FastOS}
S.~Gupta, R.~Gribonval, L.~Daudet, and Ivan Dokmani'c.
\newblock Fast optical system identification by numerical interferometry.
\newblock {\em ICASSP 2020 - 2020 IEEE International Conference on Acoustics,
  Speech and Signal Processing (ICASSP)}, pages 1474--1478, 2020.

\bibitem{Launay2020DirectFA}
Julien Launay, Iacopo Poli, Franccois Boniface, and F.~Krzakala.
\newblock Direct feedback alignment scales to modern deep learning tasks and
  architectures.
\newblock {\em ArXiv}, abs/2006.12878, 2020.

\bibitem{refinetti2020dynamics}
Maria Refinetti, St{\'e}phane d'Ascoli, Ruben Ohana, and Sebastian Goldt.
\newblock The dynamics of learning with feedback alignment.
\newblock {\em arXiv preprint arXiv:2011.12428}, 2020.

\bibitem{Akrout2019OnTA}
Mohamed Akrout.
\newblock On the adversarial robustness of neural networks without weight
  transport.
\newblock {\em ArXiv}, abs/1908.03560, 2019.

\bibitem{Lillicrap2014RandomFW}
T.~Lillicrap, D.~Cownden, D.~Tweed, and C.~Akerman.
\newblock Random feedback weights support learning in deep neural networks.
\newblock {\em ArXiv}, abs/1411.0247, 2014.

\bibitem{Krizhevsky2009LearningML}
A.~Krizhevsky.
\newblock Learning multiple layers of features from tiny images.
\newblock In {\em Learning Multiple Layers of Features from Tiny Images}, 2009.

\bibitem{kingma2014adam}
Diederik~P Kingma and Jimmy Ba.
\newblock Adam: A method for stochastic optimization.
\newblock {\em arXiv preprint arXiv:1412.6980}, 2014.

\bibitem{He2016IdentityMI}
Kaiming He, X.~Zhang, Shaoqing Ren, and Jian Sun.
\newblock Identity mappings in deep residual networks.
\newblock {\em ArXiv}, abs/1603.05027, 2016.

\bibitem{Gupta2019DontTI}
S.~Gupta, R.~Gribonval, L.~Daudet, and Ivan Dokmani{\'c}.
\newblock Don't take it lightly: Phasing optical random projections with
  unknown operators.
\newblock In {\em NeurIPS}, 2019.

\bibitem{Goibert2019AdversarialRV}
Morgane Goibert and Elvis Dohmatob.
\newblock Adversarial robustness via adversarial label-smoothing.
\newblock {\em ArXiv}, abs/1906.11567, 2019.

\bibitem{Hooker2020TheHL}
Sara Hooker.
\newblock The hardware lottery.
\newblock {\em ArXiv}, abs/2009.06489, 2020.

\end{thebibliography}

\end{document}